\documentclass{article} 
\usepackage{iclr2017_conference,times}
\usepackage{url}
\usepackage{graphicx}

\usepackage{preamble}

\title{Fast Training of Convolutional \\
Neural Networks via Kernel Rescaling}

\author{Pedro Porto Buarque de Gusm\~{a}o \& Enrico Magli \\
Department of Electronics and Telecommunications\\
Politecnico di Torino\\
Cso. Duca degli Abruzzi 24, Turin, Italy\\
\texttt{\{pedro.gusmao,enrico.magli\}@polito.it} \\
\AND
Gianluca Francini \& Skjalg Leps\o{}y \\
Joint Open Lab VISIBLE - TIM \\ 
Cso. Duca degli Abruzzi 24, Turin, Italy \\
\texttt{\{gianluca.francini,skjalg.lepsoy\}@telecomitalia.it} 
}

\begin{document}

\maketitle

\begin{abstract}

Training deep Convolutional Neural Networks (CNN) is a time consuming task that may take weeks to complete. In this article we propose a novel, theoretically founded method for reducing CNN training time without incurring any loss in accuracy. The basic idea is to begin training with a \emph{pre-train} network using lower-resolution kernels and input images, and then refine the results at the full resolution by exploiting the spatial scaling property of convolutions. We apply our method to the ImageNet winner OverFeat and to the more recent ResNet architecture and show a reduction in training time of nearly $20\%$ while test set accuracy is preserved in both cases.

\end{abstract}

\section{Introduction}\label{intro}

In the past few years, deep Convolutional Neural Networks (CNN) \citep{lecun1998gradient} have become the ubiquitous tool for solving computer vision problems such as object detection, image classification, and image segmentation \citep{ren2015faster}. Such success can be traced back to the work of \cite{Krizhevsky2012} whose 8-layer CNN  won the 2012 ImageNet Large Scale Visual Recognition Challenge (ILSVRC), showing that  multi-layer architectures are able to capture the large variability present in real world data.

Works by \cite{Simonyan2015} have shown that increasing the number of layers will consistently improve classification accuracy in this same task.  This has led to the proposal of new architectural improvements \citep{szegedy2015going,he2015deep} that allowed network depth to increase from a few \citep{Krizhevsky2012} to hundreds of layers\citep{he2015deep}. However, increase in a network's depth comes at the price of longer training times \citep{Glorot2010} mainly caused by the computationally intensive convolution operations. 

In this paper, we show that the overall training time of a \emph{target} CNN architecture can be reduced by exploiting the spatial scaling property of convolutions during early stages of learning. This is done by first training a \emph{pre-train} CNN of smaller kernel resolutions for a few epochs, followed by properly rescaling its kernels to the \emph{target}'s original dimensions and continuing training at full resolution. 

Moreover, by rescaling the kernels at different epochs, we identify a trade-off between total training time and maximum obtainable accuracy. Finally, we propose a method for choosing when to rescale kernels and evaluate our approach on recent architectures showing savings in training times of nearly $20\%$ while test set accuracy is preserved.

\section{Related Work}\label{sec:relatedworks}

Different attempts to reduce CNN training time using the standard back-propagation technique \citep{rumelhart1986learning} have been proposed in the literature. 

Regarding how convolutions are implemented, architectures with large convolution kernels \citep{Krizhevsky2012,Sermanet2013} have benefited from the Fast Fourier Transform (FFT) algorithm and reduced the number of multiplications in each 2D convolution \citep{Mathieu2014}; while the current preference for smaller $3\times{}3$ kernels has revived the interest for minimal filtering algorithms \citep{winograd1980arithmetic} as seen in recent, unpublished works by Levin et al.

Very recently, \cite{Ioffe2015} were able to reduce the total number of iterations (epochs) required to fully train a network using technique called Batch Normalization. Authors were able to reduce the internal covariance shift, inherently present in the back-propagation technique, by applying mean ad variance normalization at each layer's input.

Most relevant to our approach is the work of \cite{Chen2015} who suggested the use of function-preserving transformations to train deeper (more layers) and wider (more channels) networks  starting from shallower and narrower ones. Our approach, on the other hand, relies on scaling the spatial dimensions of convolution kernels and input images. This allows us to keep the levels of representation that are usually associated with the number layers and kernels per layer.

Finally, it is worth mentioning that much of the effort towards speeding up CNNs has been focused on inference only. Works by \cite{Lebedev2015, jaderberg2014speeding, denton2014exploiting} have used low-rank approximation to greatly reduce computational complexity of CNNs. These approaches, however, still require a network to be fully trained and they can all profit from our approach.  



\section{Proposed Method}\label{sec:method}
This section describes the rationale behind spatially scaling kernels to speed up the overall CNN training procedure.  

\begin{figure}
\centering
	\begin{tikzpicture}
	

	\node (shoe) 		at (2.6,3.9){\includegraphics[scale=0.35,frame]{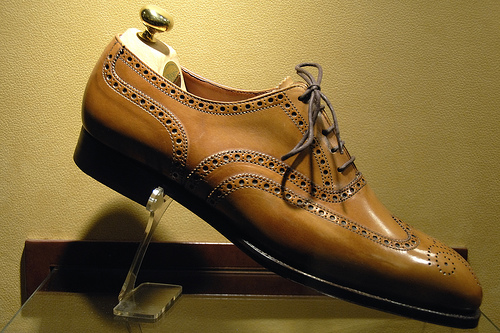}};
	\node (train) 		at (2.8,3.7){\includegraphics[scale=0.08,frame]{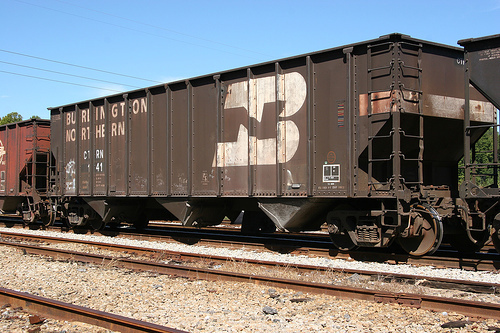}};
	\node (squirrel) 	at (3.0,3.5){\includegraphics[scale=0.08,frame]{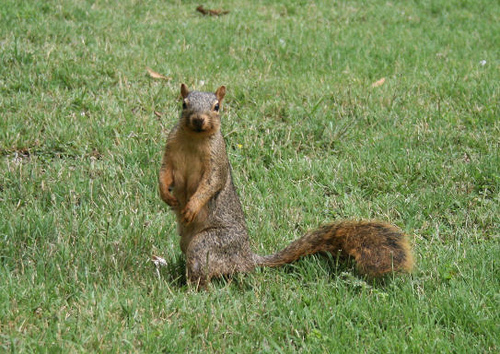}};
	
	\draw[->] (4.1,3.5) -- node [text width=1.8cm, midway,above, align=center] {Input $\left(147\times147\right)$} node [text width=1.8cm, midway,below, align=center] {$I_{\left(x,y\right)}$} (5.7,3.5);
	
	\node (randweight3) at (6.4,3.9){\includegraphics[scale=3,frame]{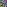}};
	\node (randweight2) at (6.6,3.7){\includegraphics[scale=3,frame]{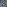}};
	\node (randweight)  at (6.8,3.5){\includegraphics[scale=3,frame]{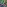}};
	
	\draw[->] (7.5,3.5) -- node [text width=1.2cm, midway, above, align=center] {Training} node [text width=1.8cm, midway,below, align=center] {$W_{\left(x,y\right)}$} (8.9, 3.5);	

	\node (weight18c) 	at (9.6,3.9){\includegraphics[scale=3,frame]{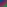}};
	\node (weight18b) 	at (9.8,3.7){\includegraphics[scale=3,frame]{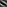}};
	\node (weight18a) 	at (10.0,3.5){\includegraphics[scale=3,frame]{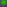}};	
	
	\node (boxpretrain)[draw=red!20, fit= (randweight3) (randweight) (weight18a)] {};
	\node [below left, color=red!70] at (boxpretrain.south east) {Pre-Train};
	
	\draw[dashed, color=black!30, very thin] (0,2.3) -- node[below, very near start, align=left] {Continue training} node[above, very near start, align=left] {Start training} (10.8, 2.3);
	
	\node (bigshoe) 	 at (2.7,0.9) {\includegraphics[scale=0.44,frame]{Fig1.png}};
	\node (bigtrain) 	 at (2.9,0.7) {\includegraphics[scale=0.10,frame]{Fig2.png}};
	\node (bigsquirrel) at (3.1,0.5) {\includegraphics[scale=0.10,frame]{Fig3.png}};
	
	\draw[->] (4.1,0.7) -- node [text width=1.8cm, midway,above, align=center] {Input $\left(231\times{}231\right)$}  node [text width=1.8cm, midway,below, align=center] {$I_{\left(sx,sy\right)}$} (5.7,0.7);
	
	
	\node  (resizedc) at (6.6,0.9){\includegraphics[scale=3,frame]{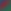}};
	\node  (resizedb) at (6.8,0.7){\includegraphics[scale=3,frame]{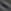}};
	\node  (resizeda) at (7.0,0.5){\includegraphics[scale=3,frame]{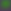}};
	
	\draw[->] (7.9,0.6) -- node [text width=1.2cm, midway, above, align=center] {Training} node [text width=1.8cm, midway,below, align=center] {$s^2W_{\left(sx,sy\right)}$}(8.9, 0.6);	
	
	\node  (retrainedc) at (9.7,0.9){\includegraphics[scale=3,frame]{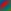}};
	\node  (retrainedb) at (9.9,0.7){\includegraphics[scale=3,frame]{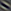}};
	\node  (retraineda) at (10.1,0.5){\includegraphics[scale=3,frame]{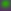}};

	\node (boxtarget)[draw=red!20, fit= (resizedc) (resizeda) (retraineda), align=right] {};	
	\node [below left, color=red!70] at (boxtarget.south east) {Target};
	
	\draw[<-,postaction={decorate,decoration={text along path,text align = center,text={Upscale}}}] (resizedb.north east) to node[swap] {}  (weight18b.south west);
	
	\end{tikzpicture}
	\caption{Training starts with a \emph{pre-train} network of smaller convolution kernels and input images. After a number of epochs, kernels are resized to the \emph{target}'s resolution and training continues as scheduled.}
	\label{fig:trainingchain}
\end{figure}
\subsection{Spatially Scaling Convolutions}

The time scaling property of convolutions states that convolution between two time-scaled signals $I\left(sx\right)$ and $W\left(sx\right)$ can be obtained by time-scaling the result of convolving the original inputs $I\left(x\right)$ and $W\left(x\right)$, followed by an amplitude-scaling of $1/{\left|s\right|}$.

This property can be extended to continuous 2D signals (Equations \ref{eq:convoriginal} and \ref{eq:convscaled}) where in this case it is better denoted as the \emph{spatial scaling} property. 

\begin{align}
	I(x,y)*W(x,y) = & Y(x,y) \label{eq:convoriginal}\\
	I(sx,sy)*W(sx,sy) = & \frac{1}{s^2}Y(sx,sy) \label{eq:convscaled}
\end{align}

If applied to the context of CNNs, this property would suggest that the output of a convolutional layer could also be obtained from the spatially downsized versions of both the layer's input and its convolution kernels. 

Benefits of this possibility can be seen in Equation \ref{eq:nummult}, which represents the number of multiplications performed by a convolutional layer $l$, having $C_{l-1}$ and $C_{l}$ input and output channels, when both $h_l\times{}h_l$ input and $k_l\times{}k_l$ convolution kernels are spatially scaled by a factor $s_l$.

\begin{equation}\label{eq:nummult}
	M\left(l,s_l\right) = C_{l-1}\left(\frac{k_l}{s_l}\right)^2\left(\frac{h_l}{s_l}-\frac{k_l}{s_l}+1\right)^2C_{l}
\end{equation}


When compared to its unscaled version, i.e.\ $s_l=1$, one can establish the bounds in Equation \ref{eq:bounds} by considering both extremes  $h_l = k_l$  and $\left(h_l-k_l\right) \gg s_l$, which in turn guarantees a minimum reduction in the number of multiplications proportional to $1/\left|s_l^2\right|$.

\begin{equation}\label{eq:bounds}
 M\left(l,1\right)/s_l^4 <  M\left(l,s_l\right) \leq  M\left(l,1\right)/s_l^2
\end{equation}

The spatial scaling property is, of course, valid only in the continuous domain. Working with downsized versions of inputs will usually result in irreversible loss of spatial resolution and accuracy. 
However, as shown in the following sections, for moderate values of $s_l$ this property can still be exploited during early stages of training,  where the network is still learning the basic structures for its kernels. This can be done by first training an otherwise identical network of smaller kernel resolutions, followed by an upscaling to the target kernel resolution and continuing training.

\subsection{Pre-training Setup}
\label{subsec:pretraining}

During the pre-training phase, a \emph{pre-train} network of architecture similar to the \emph{target} one having downscaled kernel resolutions shall be trained. Equation \ref{eq:nummult} guarantees that during this phase the training process will run faster. 

Generating this \emph{pre-train} network from a \emph{target} network architecture requires choosing new spatial resolutions for each convolutional layer as well as making the necessary adjustments so that fully-connected layers will have compatible input-output dimensions.  

\subsubsection{Convolution Kernels}
\label{subsubsec:convolutionalkernels1}

When deciding on the new kernel resolutions, a trade-off between speed and accuracy must be considered.

Selecting kernels much smaller than the originals will cripple the layer's ability to extract and forward high-frequency information, while too conservative downscaling will lead to insignificant improvements in training speeds. Therefore, we suggest the use of Table \ref{table:conversiontable} for choosing the pre-train kernel resolution given a target kernel resolution. We have found those values to provide a good compromise between these factors after the complete training process.

\begin{table}
\caption{Suggested kernel resolution conversions with relative resize factors and bounds.}
\label{table:conversiontable}
\centering
\begin{tabular}{ccccccc}
\hline\noalign{\smallskip}
 Target & & Pre-train & & $s_l$ & $s_{l}^2$ & $s_{l}^4$\\ 
\hline\noalign{\smallskip}
$1\times{}1$		& & $1\times{}1$ & & $1.00$ & $1.00$ & $1.00$ \\
$3\times{}3$		& & $2\times{}2$ & & $1.50$ & $2.25$ & $5.06$ \\
$5\times{}5$		& & $3\times{}3$ & & $1.67$ & $2.77$ & $7.71$ \\
$7\times{}7$		& & $5\times{}5$ & & $1.40$ & $1.96$ & $3.84$  \\
$11\times{}11$		& & $7\times{}7$ & & $1.57$ & $2.49$ & $6.09$ \\
\hline
\end{tabular}
\end{table}


 

 
Once the new kernel sizes have been chosen, it is necessary to adjust the network's internal parameters so that each convolutional layer closely satisfies Equation \ref{eq:convscaled}. In order to do so, it is important to observe that a CNN architecture for image classification usually reflects two distinct stages of processing. The first stage contains various layers of convolution and pooling that act as feature extractors. They output \emph{feature-maps} whose spatial resolution depends on that layer's input resolution. The second stage, on the other hand, acts as a classifier and can be identified by the presence of \emph{fully-connected} layers of fixed input and output dimensions. 

Solving the input-output dependencies present in the feature extraction phase should be done starting from the input image itself since it has no constraints with any previous layer. Input images also have large spatial resolution so that it should be straightforward to choose a smaller integer length whose ratio with respect to the original image will closely approximate the first chosen scaling factor $s_1$.

Upper layers will generally not have such flexibility since the following feature-maps will usually have lower spatial resolution. For these layers, the input-output scaling requirement are met by spatially padding or cropping the incoming feature-map.

\subsubsection{Fully-connected Interface} 
\label{subsubsec:fullyconnect1}

Special attention must be paid to the interface between convolutional layers and fully-connected ones since the latter require fixed input sizes. 

Activations in a fully-connected layer can be represented as a matrix-vector multiplication where each column in the weight matrix is associated to a particular neuron and the number of rows defines the layer's input size. In this interpretation, before entering a fully-connected layer, feature-maps having $C$ channels and spatial resolution of $W\times{H}$ must be reshaped into a vector representation  $\mathbf{x} \in \mathbb{R}^{CHW}$.

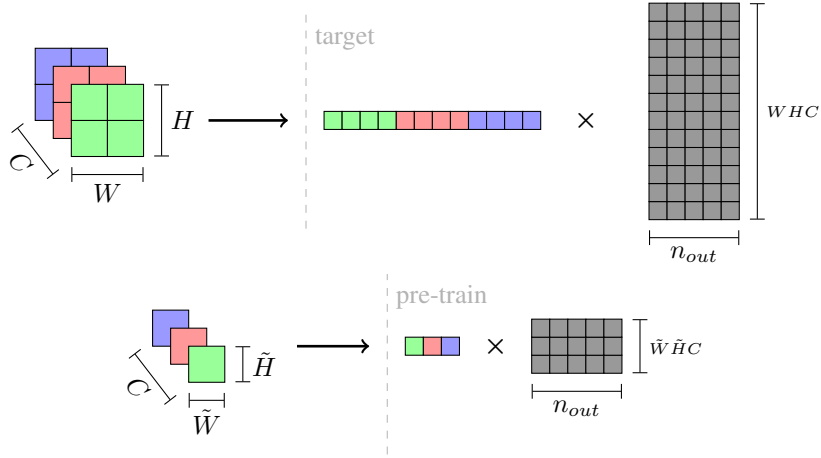
\begin{figure}
\centering
	\begin{tikzpicture}[scale=1.2]
	\pgfmathsetmacro{\step}{0.4};
	\pgfmathsetmacro{\stepx}{\step};
	\pgfmathsetmacro{\stepy}{\step};
	\pgfmathsetmacro{\spacebetweenmatrices}{0.5*\step};
	\pgfmathsetmacro{\vectorstep}{0.2};
	\pgfmathsetmacro{\numberoutput}{5}
	
	\pgfmathsetmacro{\startx}{0};
	\pgfmathsetmacro{\sizex}{2};
	\pgfmathsetmacro{\starty}{0};
	\pgfmathsetmacro{\sizey}{2};
	\pgfmathsetmacro{\vectorstartx}{\startx + 8*\stepx};
    \pgfmathsetmacro{\vectorstarty}{\starty -0.25*\stepy};
    \pgfmathsetmacro{\vectorstartmatrixx}{\vectorstartx + 9*\stepx}
    \pgfmathsetmacro{\vectorstartmatrixy}{\vectorstarty - 2.5*\stepy}
    
    \foreach \x in {1,...,\sizex}
    	\foreach \y in {1,...,\sizey}
    		\filldraw[fill=blue!40!white, draw=black] (\startx + \x*\stepx -\stepx, \starty + \y*\stepy -\stepy) rectangle (\startx + \x*\stepx, \starty + \y*\stepy ); 

	\pgfmathsetmacro{\startx}{\startx + \spacebetweenmatrices};
	\pgfmathsetmacro{\starty}{\starty -\spacebetweenmatrices};
		
    \foreach \x in {1,...,\sizex}
    	\foreach \y in {1,...,\sizey}
    		\filldraw[fill=red!40!white, draw=black] (\startx + \x*\stepx -\stepx, \starty + \y*\stepy -\stepy) rectangle (\startx + \x*\stepx, \starty + \y*\stepy );     		
    
	\pgfmathsetmacro{\startx}{\startx + \spacebetweenmatrices};	
	\pgfmathsetmacro{\starty}{\starty - \spacebetweenmatrices};
    \foreach \x in {1,...,\sizex}
    	\foreach \y in {1,...,\sizey}
    		\filldraw[fill=green!40!white, draw=black] (\startx + \x*\stepx -\stepx, \starty + \y*\stepy -\stepy) rectangle (\startx + \x*\stepx, \starty + \y*\stepy );     	
    			
	 \draw[|-|] (\startx + -1.5*\stepx, \starty + 0.8*\stepy) -- node [text width=1.8cm, sloped, anchor=center, below, align=center] {$C$} (\startx + -0.5*\stepx, \starty + -0.5*\stepy);
	 \draw[|-|] (\startx, \starty + -0.5*\stepy) -- node [text width=1.8cm, sloped, anchor=center, below, align=center] {$W$} (\startx + \sizex*\stepx,\starty + -0.5*\stepy);
	 \draw[|-|] (\startx + \sizex*\stepx + 0.5*\step, \starty) -- node [text width=1.8cm, right] {$H$} (\startx + \sizex*\stepx + 0.5*\step, \starty + \sizey*\stepy);
    
    \draw[->,line width=1pt] (\startx + 3.8*\stepx, \starty + \stepy)  --  (\startx + 6*\stepx, \starty +\stepy);
    
    \draw[dashed, color=black!30, very thin] (\startx + 6.5*\stepx, \starty - 2*\stepy) -- node[very near end, right]{target} (\startx + 6.5*\stepx, \starty + 4*\stepy);
    
    \foreach \x in {1,...,4}
    		\filldraw[fill=green!40!white, draw=black] (\vectorstartx + \x*\vectorstep - \vectorstep, \vectorstarty) rectangle (\vectorstartx + \x*\vectorstep, \vectorstarty + \vectorstep ); 
    
    \pgfmathsetmacro{\vectorstartx}{\vectorstartx + \sizex*\sizey*\vectorstep};
    \foreach \x in {1,...,4}
    		\filldraw[fill=red!40!white, draw=black] (\vectorstartx + \x*\vectorstep - \vectorstep, \vectorstarty) rectangle (\vectorstartx + \x*\vectorstep, \vectorstarty + \vectorstep ); 

    \pgfmathsetmacro{\vectorstartx}{\vectorstartx + \sizex*\sizey*\vectorstep};
    \foreach \x in {1,...,4}
    		\filldraw[fill=blue!40!white, draw=black] (\vectorstartx + \x*\vectorstep - \vectorstep, \vectorstarty) rectangle (\vectorstartx + \x*\vectorstep, \vectorstarty + \vectorstep ); 
    
    \node at (\vectorstartx+4*\vectorstep + 0.5, \vectorstarty + 0.1) {$\boldsymbol{\times{}}$}; 
        		
    \foreach \x in {1,...,\numberoutput}
    	\foreach \y in {1,...,12}
    		\filldraw[fill=black!40!white, draw=black] (\vectorstartmatrixx + \x*\vectorstep -\vectorstep, \vectorstartmatrixy + \y*\vectorstep -\vectorstep) rectangle (\vectorstartmatrixx + \x*\vectorstep, \vectorstartmatrixy + \y*\vectorstep );
    
    \draw[|-|] (\vectorstartmatrixx + \numberoutput*\vectorstep + 0.2 , \vectorstartmatrixy ) -- node [near end, right, midway] {\tiny $WHC$} ( \vectorstartmatrixx + \numberoutput*\vectorstep + 0.2, \vectorstartmatrixy + 3*4*\vectorstep );
    
    \draw[|-|] (\vectorstartmatrixx, \vectorstartmatrixy -0.2) -- node [below, align=center] {$n_{out}$} (\vectorstartmatrixx + \numberoutput*\vectorstep, \vectorstartmatrixy -0.2);
    		
    \end{tikzpicture}
    
    \begin{tikzpicture}[scale=1.20]
	\pgfmathsetmacro{\step}{0.4};
	\pgfmathsetmacro{\stepx}{\step};
	\pgfmathsetmacro{\stepy}{\step};
	\pgfmathsetmacro{\spacebetweenmatrices}{0.5*\step};
	\pgfmathsetmacro{\vectorstep}{0.2};
	\pgfmathsetmacro{\numberoutput}{5}
    \pgfmathsetmacro{\startx}{20*\stepx};
    \pgfmathsetmacro{\sizex}{1};
	\pgfmathsetmacro{\starty}{0};
	\pgfmathsetmacro{\sizey}{1};
	
    \pgfmathsetmacro{\vectorstartx}{\startx + 8*\stepx};
    \pgfmathsetmacro{\vectorstarty}{\starty -0.25*\stepy};
    \pgfmathsetmacro{\vectorstartmatrixx}{\vectorstartx + 2.5*\stepx}
    \pgfmathsetmacro{\vectorstartmatrixy}{\vectorstarty - 0.5*\stepy}
    
    \foreach \x in {1,...,\sizex}
    	\foreach \y in {1,...,\sizey}
    		\filldraw[fill=blue!40!white, draw=black] (\startx + \x*\stepx -\stepx, \starty + \y*\stepy -\stepy) rectangle (\startx + \x*\stepx, \starty + \y*\stepy ); 

	\pgfmathsetmacro{\startx}{\startx + \spacebetweenmatrices};
	\pgfmathsetmacro{\starty}{\starty -\spacebetweenmatrices};
		
    \foreach \x in {1,...,\sizex}
    	\foreach \y in {1,...,\sizey}
    		\filldraw[fill=red!40!white, draw=black] (\startx + \x*\stepx -\stepx, \starty + \y*\stepy -\stepy) rectangle (\startx + \x*\stepx, \starty + \y*\stepy );     		
    
	\pgfmathsetmacro{\startx}{\startx + \spacebetweenmatrices};	
	\pgfmathsetmacro{\starty}{\starty - \spacebetweenmatrices};
    \foreach \x in {1,...,\sizex}
    	\foreach \y in {1,...,\sizey}
    		\filldraw[fill=green!40!white, draw=black] (\startx + \x*\stepx -\stepx, \starty + \y*\stepy -\stepy) rectangle (\startx + \x*\stepx, \starty + \y*\stepy );     	
    			
	 \draw[|-|] (\startx + -1.5*\stepx, \starty + 0.8*\stepy) -- node [text width=1.8cm, sloped, anchor=center, below, align=center] {$C$} (\startx + -0.5*\stepx, \starty + -0.5*\stepy);
	 \draw[|-|] (\startx, \starty + -0.5*\stepy) -- node [text width=1.8cm, sloped, anchor=center, below, align=center] {$\tilde{W}$} (\startx + \sizex*\stepx,\starty + -0.5*\stepy);
	 \draw[|-|] (\startx + \sizex*\stepx + 0.5*\step, \starty) -- node [text width=1.8cm, right] {$\tilde{H}$} (\startx + \sizex*\stepx + 0.5*\step, \starty + \sizey*\stepy);
    
    \draw[->,line width=1pt] (\startx + 3.0*\stepx, \starty + \stepy)  --  (\startx + 5*\stepx, \starty +\stepy);
    
    \draw[dashed, color=black!30, very thin] (\startx + 5.5*\stepx, \starty - 2*\stepy) -- node[very near end, right]{pre-train} (\startx + 5.5*\stepx, \starty + 3*\stepy);
    
    \foreach \x in {1,...,1}
    		\filldraw[fill=green!40!white, draw=black] (\vectorstartx + \x*\vectorstep - \vectorstep -0.4, \vectorstarty) rectangle (\vectorstartx + \x*\vectorstep -0.4, \vectorstarty + \vectorstep ); 
    
    \pgfmathsetmacro{\vectorstartx}{\vectorstartx + \sizex*\sizey*\vectorstep};
    \foreach \x in {1,...,1}
    		\filldraw[fill=red!40!white, draw=black] (\vectorstartx + \x*\vectorstep - \vectorstep-0.4, \vectorstarty) rectangle (\vectorstartx + \x*\vectorstep -0.4, \vectorstarty + \vectorstep ); 

    \pgfmathsetmacro{\vectorstartx}{\vectorstartx + \sizex*\sizey*\vectorstep};
    \foreach \x in {1,...,1}
    		\filldraw[fill=blue!40!white, draw=black] (\vectorstartx + \x*\vectorstep - \vectorstep-0.4, \vectorstarty) rectangle (\vectorstartx + \x*\vectorstep-0.4, \vectorstarty + \vectorstep ); 
    
    \node at (\vectorstartx+ \vectorstep , \vectorstarty + 0.1) {$\boldsymbol{\times{}}$}; 
        		
    \foreach \x in {1,...,\numberoutput}
    	\foreach \y in {1,...,3}
    		\filldraw[fill=black!40!white, draw=black] (\vectorstartmatrixx + \x*\vectorstep -\vectorstep, \vectorstartmatrixy + \y*\vectorstep -\vectorstep) rectangle (\vectorstartmatrixx + \x*\vectorstep, \vectorstartmatrixy + \y*\vectorstep );
    
    \draw[|-|] (\vectorstartmatrixx + \numberoutput*\vectorstep +0.2 , \vectorstartmatrixy ) -- node [near end, right, midway] {\tiny $\tilde{W}\tilde{H}C$} ( \vectorstartmatrixx + \numberoutput*\vectorstep +0.2, \vectorstartmatrixy + 3*\vectorstep );
    
    \draw[|-|] (\vectorstartmatrixx, \vectorstartmatrixy -0.2) -- node [below, align=center] {$n_{out}$} (\vectorstartmatrixx + \numberoutput*\vectorstep, \vectorstartmatrixy -0.2);

	\end{tikzpicture}
	\caption{Visual representation of the interface between convolutional and fully-connected layers. Feature-maps from a convolutional layer are first vectorized before entering a fully-connected layer, whose weights are usually represented in matrix form. The number of input must be selected according to the new feature-map spatial resolution ($\tilde{W}, \tilde{H}$) and the number of output neurons $n_{out}$ is kept invariant.}
	\label{fig:interfacefullyconnected}
\end{figure}

Since the new spatial dimensions in the \emph{pre-train} architecture will produce feature-maps of smaller resolutions $\tilde{W}\times{}\tilde{H}$, the weight matrix in the fully-connected layer must be adapted accordingly. In our representation, this means that the number of inputs (rows) in the weight matrix shall be reduced from $CHW$ to $C\tilde{H}\tilde{W}$, while the number of output neurons (columns) $n_{out}$ is kept invariant. A visual representation of this procedure is seen in Figure \ref{fig:interfacefullyconnected}.

Subsequent layers should need no further modification and the \emph{pre-train} network can be trained until a given stopping criterion is met, e.g.\ classification accuracy on a validation set starts to plateau. 


\subsection{Resizing and Continuing Training}
\label{subsec:transferringkernels}

Once the stopping criterion has been reached by the \emph{pre-train} network, its structure must be modified back to the original \emph{target} network. 

\subsubsection{Convolution Kernels} 
\label{subsubsec:convolutionalkernels2}

As seen in Equation \ref{eq:convscaled}, just spatially resizing both input and kernels would result in an amplitude scaled version of the expected convolution, meaning that the scaling factor would propagate to all subsequent layers. This problem can be avoided simply by scaling the amplitude of the resized kernels by $s_l^2$. That is, given a pre-trained kernel matrix $W_{l,c,1}\left(x,y\right)$ that represents the weights of a convolution kernel $c$ from layer $l$, the corresponding weights to be used in the target network are $W_{l,c,s_l}\left(x,y\right) = s_l^2W_{l,c,1}\left(sx,sy\right)$ so that the amplitude gain caused by the convolution operation cancels out.

Moreover, associated to each convolution operation is a bias component $b_{l,c,1}$ that need not be scaled since it has a constant value. This resizing procedure should be carried out for every channel $c$ in every convolutional layer $l$ as described in Equations \ref{eq:convkernel} and \ref{eq:convbias}.

\begin{align}
W_{l,c,s_l}(x,y) &= s_l^2W_{l,c,1}(s_lx,s_ly) \label{eq:convkernel} \\
b_{l,c,s_l}(x,y) &= b_{l,c,1}(s_lx,s_ly) \label{eq:convbias}
\end{align}

Kernels in our experiments were spatially upscaled using bilinear interpolation.  Although other interpolation methods were tested, \emph{pre-train} kernel resolutions were too small to benefit from higher order interpolation such as bicubic.



\subsubsection{Fully-connected Interface} 
\label{subsubsec:fullyconnect2}

Again, special attention should be paid to the interface between convolutional and fully-connected layers. 

According to the usual interpretation described in \ref{subsubsec:fullyconnect1}, the output of a convolutional layer must be vectorized before serving as input to a fully-connected layer, implying a loss of its explicit spatial representation. However, since the incoming feature-maps do contain intra-channel correlation, such information is still present and it is captured by the weights of the fully-connected layer.

\begin{figure}
	\centering
	\begin{tikzpicture}[scale=1.20]
		\pgfmathsetmacro{\step}{0.4};
		\pgfmathsetmacro{\stepx}{\step};
		\pgfmathsetmacro{\stepy}{\step};
		\pgfmathsetmacro{\spacebetweenmatrices}{0.5*\step};
		\pgfmathsetmacro{\vectorstep}{0.2};
		\pgfmathsetmacro{\vectorstepx}{\vectorstep};
		\pgfmathsetmacro{\vectorstepy}{\vectorstep};
		\pgfmathsetmacro{\numberoutput}{5}
			    
		\pgfmathsetmacro{\startx}{0};
		\pgfmathsetmacro{\sizepretrainx}{1};
		\pgfmathsetmacro{\sizetargetx}{2};
		\pgfmathsetmacro{\starty}{0};
		\pgfmathsetmacro{\sizepretrainy}{1};	
		\pgfmathsetmacro{\sizetargety}{2};
				
		\pgfmathsetmacro{\vectorstartx}{\startx + 8*\stepx};
		\pgfmathsetmacro{\vectorstarty}{\starty -0.25*\stepy};
		\pgfmathsetmacro{\vectorstartmatrixx}{\vectorstartx + 2.5*\stepx}
		\pgfmathsetmacro{\vectorstartmatrixy}{\vectorstarty - 0.5*\stepy}
			    
		\draw[|-|] (-\vectorstepx, \starty) -- node [near end, left, midway] {\tiny $\tilde{W}\tilde{H}C$} (-\vectorstepx, \starty + 3*\vectorstepy);
		\draw[|-|] (\startx, \starty -1.5*\vectorstepy) -- node [below, align=center] {$n_{out}$} (\startx + \numberoutput*\vectorstepx, \starty-1.5*\vectorstepy);
		
		\foreach \y in {1,...,3}
			\filldraw[fill=black!40!white, draw=black] (\startx, \starty + \y*\vectorstepy - \vectorstepy) rectangle (\startx + \vectorstepx, \starty + \y*\vectorstepy);
		\pgfmathsetmacro{\startx}{\startx + \vectorstepx};
			     	
		\filldraw[fill=blue!40!white, draw=black] (\startx, \starty) rectangle (\startx + \vectorstepx, \starty + \vectorstepy); 
		\filldraw[fill=red!40!white, draw=black] (\startx, \starty + \vectorstepy) rectangle (\startx + \vectorstepx, \starty + 2*\vectorstepy); 
		\filldraw[fill=green!40!white, draw=black] (\startx, \starty + 2*\vectorstepy) rectangle (\startx + \vectorstepx, \starty + 3*\vectorstepy); 
		\pgfmathsetmacro{\startx}{\startx + \vectorstepx};
				      
		\foreach \x in {1,...,3} {
			\foreach \y in {1,...,3}
				\filldraw[fill=black!40!white, draw=black] (\startx + \x*\vectorstepx -\vectorstepx, \starty + \y*\vectorstepy - \vectorstepy) rectangle (\startx + \x*\vectorstepx, \starty + \y*\vectorstepy);}
		\pgfmathsetmacro{\startx}{\startx + 3*\vectorstepx};     
			
		\draw[dashed, color=black!30, very thin] (\startx + \stepx, \starty - 2*\stepy) -- node[very near end, left]{pre-train} (\startx + \stepx, \starty + 7*\stepy);
		\pgfmathsetmacro{\startx}{\startx + 2*\stepx}; 
			    
		\pgfmathsetmacro{\starty}{\starty + 0.5*\stepy};
		\filldraw[fill=blue!40!white, draw=black] (\startx, \starty) rectangle (\startx + \stepx, \starty + \stepy );
		\pgfmathsetmacro{\startx}{\startx + \spacebetweenmatrices};
		\pgfmathsetmacro{\starty}{\starty -\spacebetweenmatrices};
				
		\filldraw[fill=red!40!white, draw=black] (\startx, \starty) rectangle (\startx + \stepx, \starty + \stepy );
		\pgfmathsetmacro{\startx}{\startx + \spacebetweenmatrices};
		\pgfmathsetmacro{\starty}{\starty -\spacebetweenmatrices};  
				
		\filldraw[fill=green!40!white, draw=black] (\startx, \starty) rectangle (\startx + \stepx, \starty + \stepy );
		\pgfmathsetmacro{\startx}{\startx + 2};

		\foreach \x in {1,...,\sizetargetx}
			\foreach \y in {1,...,\sizetargety}
				\filldraw[fill=blue!40!white, draw=black] (\startx + \x*\stepx -\stepx, \starty + \y*\stepy -\stepy) rectangle (\startx + \x*\stepx, \starty + \y*\stepy ); 
			
		\pgfmathsetmacro{\startx}{\startx + \spacebetweenmatrices};
		\pgfmathsetmacro{\starty}{\starty -\spacebetweenmatrices};
					
		\foreach \x in {1,...,\sizetargetx}
			\foreach \y in {1,...,\sizetargety}
				\filldraw[fill=red!40!white, draw=black] (\startx + \x*\stepx -\stepx, \starty + \y*\stepy -\stepy) rectangle (\startx + \x*\stepx, \starty + \y*\stepy );     		
			    
		\pgfmathsetmacro{\startx}{\startx + \spacebetweenmatrices};	
		\pgfmathsetmacro{\starty}{\starty - \spacebetweenmatrices};
		
		\foreach \x in {1,...,\sizetargetx}
			\foreach \y in {1,...,\sizetargety}
				\filldraw[fill=green!40!white, draw=black] (\startx + \x*\stepx -\stepx, \starty + \y*\stepy -\stepy) rectangle (\startx + \x*\stepx, \starty + \y*\stepy );  
		\pgfmathsetmacro{\starty}{\starty - 0.5*\stepy};
			    
		\pgfmathsetmacro{\startx}{\startx + 6*\spacebetweenmatrices};
		\pgfmathsetmacro{\starty}{\starty + 4*\spacebetweenmatrices};		
		\draw[dashed, color=black!30, very thin] (\startx + \stepx, \starty - 2*\stepy) -- node[very near end, right]{target} (\startx + \stepx, \starty + 7*\stepy);
		\pgfmathsetmacro{\startx}{\startx + 2*\stepx};
		
		\pgfmathsetmacro{\starty}{\starty - 2*\stepy};
		\pgfmathsetmacro{\starty}{-1.2};
		\foreach \y in {1,...,12}
			\filldraw[fill=black!40!white, draw=black] (\startx, \starty + \y*\vectorstepy - \vectorstepy) rectangle (\startx + \vectorstepx, \starty + \y*\vectorstepy);
		\pgfmathsetmacro{\startx}{\startx + \vectorstepx};
		
		\foreach \y in {1,...,4}
			\filldraw[fill=blue!40!white, draw=black] (\startx, \starty + \y*\vectorstepy -\vectorstepy) rectangle (\startx + \vectorstepx, \starty + \y*\vectorstepy); 
		\foreach \y in {5,...,8}
			\filldraw[fill=red!40!white, draw=black] (\startx, \starty + \y*\vectorstepy -\vectorstepy) rectangle (\startx + \vectorstepx, \starty + \y*\vectorstepy); 
		\foreach \y in {9,...,12}
			\filldraw[fill=green!40!white, draw=black] (\startx, \starty + \y*\vectorstepy -\vectorstepy) rectangle (\startx + \vectorstepx, \starty + \y*\vectorstepy); 
		\pgfmathsetmacro{\startx}{\startx + \vectorstepx};	
		
		\foreach \x in {1,...,3} {
			\foreach \y in {1,...,12}
				\filldraw[fill=black!40!white, draw=black] (\startx + \x*\vectorstepx -\vectorstepx, \starty + \y*\vectorstepy - \vectorstepy) rectangle (\startx + \x*\vectorstepx, \starty + \y*\vectorstepy);}
	   
		 \draw[|-|] (\startx - 2*\vectorstep, \vectorstartmatrixy -3*\stepy) -- node [below, align=center] {$n_{out}$} (\startx + \numberoutput*\vectorstep - 2*\vectorstep, \vectorstartmatrixy -3*\stepy);
		 
		 \draw[|-|] (\vectorstartmatrixx + 9*\stepx + 0.1, \starty) -- node [near end, right, midway] {\tiny $WHC$} ( \vectorstartmatrixx + 9*\stepx + 0.1, \starty + 12*\vectorstepy);
			    	 
	 \draw[->,postaction={decorate,decoration={text along path,text align=center,text={reshape}}}](1.5*\vectorstepx, 4*\vectorstepy ) .. controls (2.75*\vectorstepx, 4*\stepy)  and (4*\vectorstepx + 3*\stepx, 4*\stepy ) .. (4*\vectorstepx + 3*\stepx, 3.5*\vectorstepy ) ;
	 
	 \draw[->,postaction={decorate,decoration={text along path,text align=center,text={upsize}}}] (5*\vectorstepx + 3*\stepx, 3.5*\vectorstepy ) .. controls (6*\vectorstepx + 3*\stepx, 4*\stepy) and (11*\stepx, 4*\stepy) .. (11*\stepx , 3.5*\vectorstepy );
	 
	 \draw[->,postaction={decorate,decoration={text along path,text align=center,text={vectorize}}}] (12*\stepx , 3.5*\vectorstepy ) .. controls (12*\stepx +\vectorstepx, 5*\stepy ) and (16*\stepx + 2*\vectorstepx, 5*\stepy) .. (16*\stepx + 2.3*\vectorstepx,  3*\stepy + 0.5*\vectorstepy);

	\end{tikzpicture}
	\caption{Rescaling weights in fully-connected layer back to \emph{target}'s dimensions. Each column in the fully-connected weight matrix is reshaped to match the \emph{pre-train} feature-map dimensions. Rescaling is applied in the same fashion as regular convolutional kernels and weights are then vectorized to the \emph{target}'s new weight matrix.}
	\label{fig:usizingfullyconnected}
\end{figure}
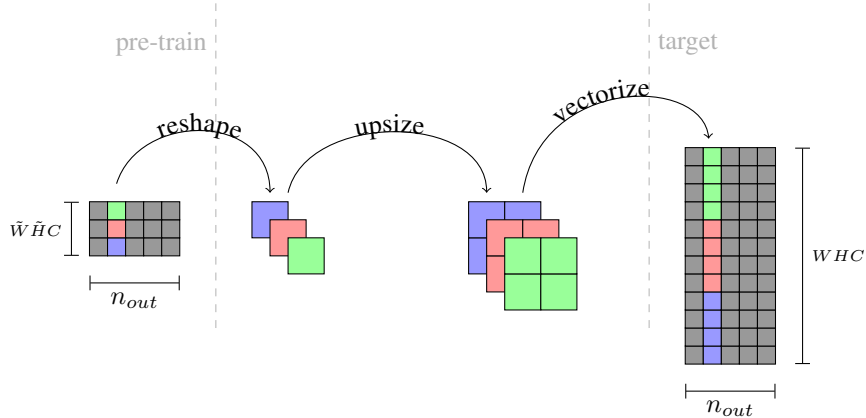
  
In order to exploit this correlation and be able to apply our method, the fully connected layer shall be reinterpreted as a convolutional layer. In other words, each column in the weight matrix must first be reshaped into a third-order tensor $\mathcal{X} \in \mathbb{R}^{C\times{}\tilde{H}\times{}\tilde{W}}$ of the same dimensions as the original incoming feature-maps so that we can apply the same resizing rule defined in Equations \ref{eq:convkernel} and \ref{eq:convbias}. This will produce a new tensor $\mathcal{Y} \in \mathbb{R}^{C\times{}H\times{}W}$ that must be then vectorized into the new weight matrix whose dimensions are consistent with the \emph{target} network. Figure \ref{fig:trainingchain} illustrates the overall training process.

Finally, weights from successive fully-connected layers must simply be copied to the target model. 


\section{Preliminary Experiments}
\label{sec:preliminary}

In order to assess the proposed approach, we must first estimate upper and lower bounds in terms of accuracy and training times set by the \emph{target} network and its \emph{pre-train} counterpart. To do this we use as baseline to our investigation the \emph{fast} variant of 2013 ImageNet localization winner OverFeat \citep{Sermanet2013}. 

The original OverFeat-\emph{fast} contains five convolutional layers followed by three fully-connected ones that classify $231\times{}231$ RGB images among the 1000 classes defined by the ImageNet dataset. By following the steps set in subsection \ref{subsec:pretraining} we generate a \emph{pre-train} model of input resolution $147\times{}147$ and do not apply padding at the last convolutional layer. 
 
During our experiments we use an Nvidia Tesla K80 GPU to train and test both networks with ImageNet 2012 CLS-LOC training and validation datasets.
We use mini-batches of 128 images and 10k mini-batches per epoch. We use an initial learning rate of \num{1e-2} and lower it to \num{5e-3}, \num{1e-3}, \num{5e-4}, and \num{1e-4} at the end of epochs 18, 29, 43, and 52 respectively, until epoch 65 when training is halted. A weight decay of \num{1e-4} is also applied until the end of epoch 29 and a momentum of $0.9$ is used during the entire training. For both networks, weights in each layer are initialized uniformly at random  in the interval $\left[-\frac{1}{\sqrt{n}},\frac{1}{\sqrt{n}}\right]$, where $n$ is the layer's number of weights.  

\begin{figure}
	\centering
	\includegraphics[width=0.70\linewidth]{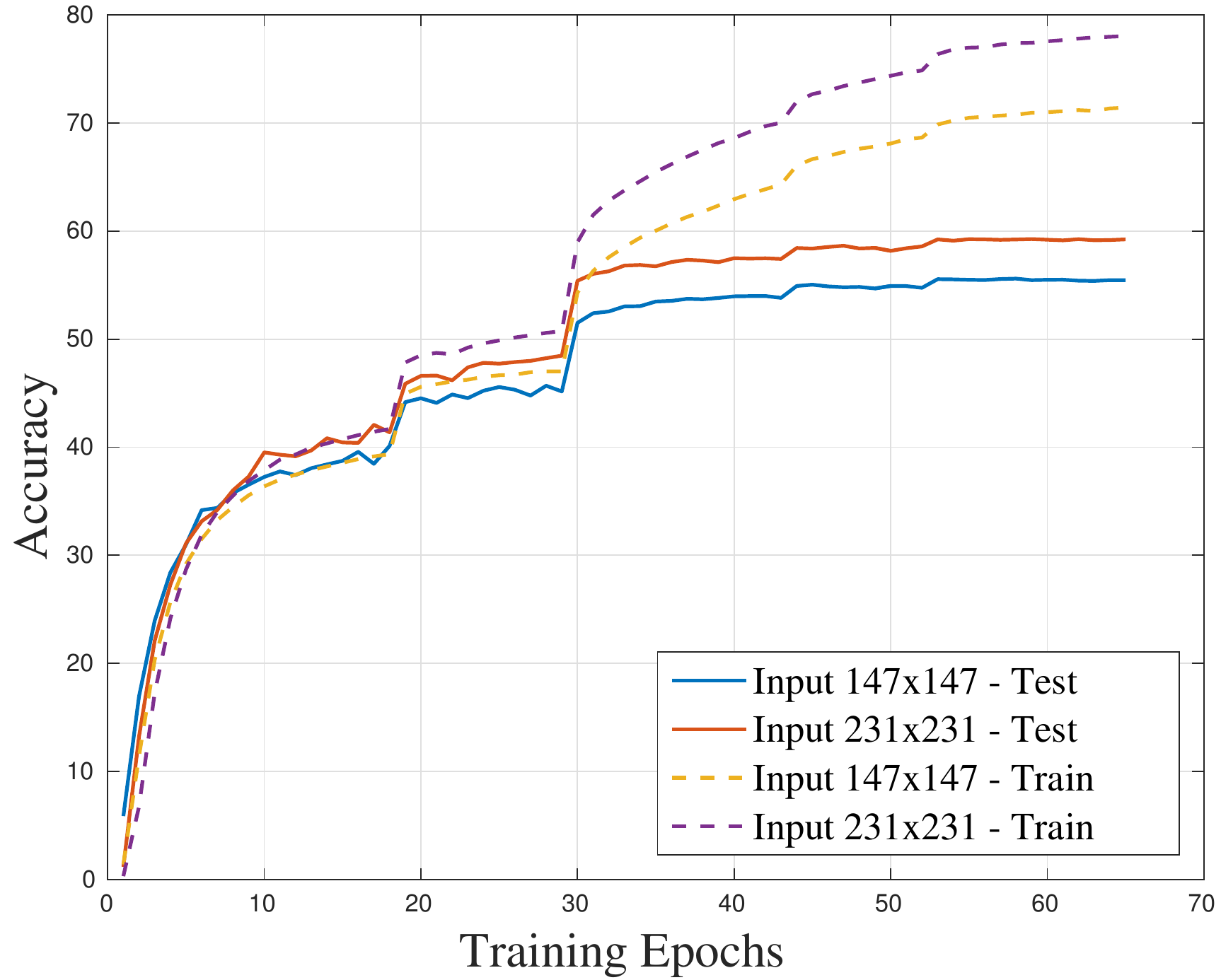}
	\caption{Accuracy as function of epochs obtained using both original OverFeat-\emph{fast} of input resolution $231\times{}231$ and its \emph{pre-train} counterpart having $147\times{}147$ input resolution.}
	\label{fig:originalvsdownsizedepochs}
\end{figure}

\begin{figure}
	\centering
	\includegraphics[width=0.70\columnwidth]{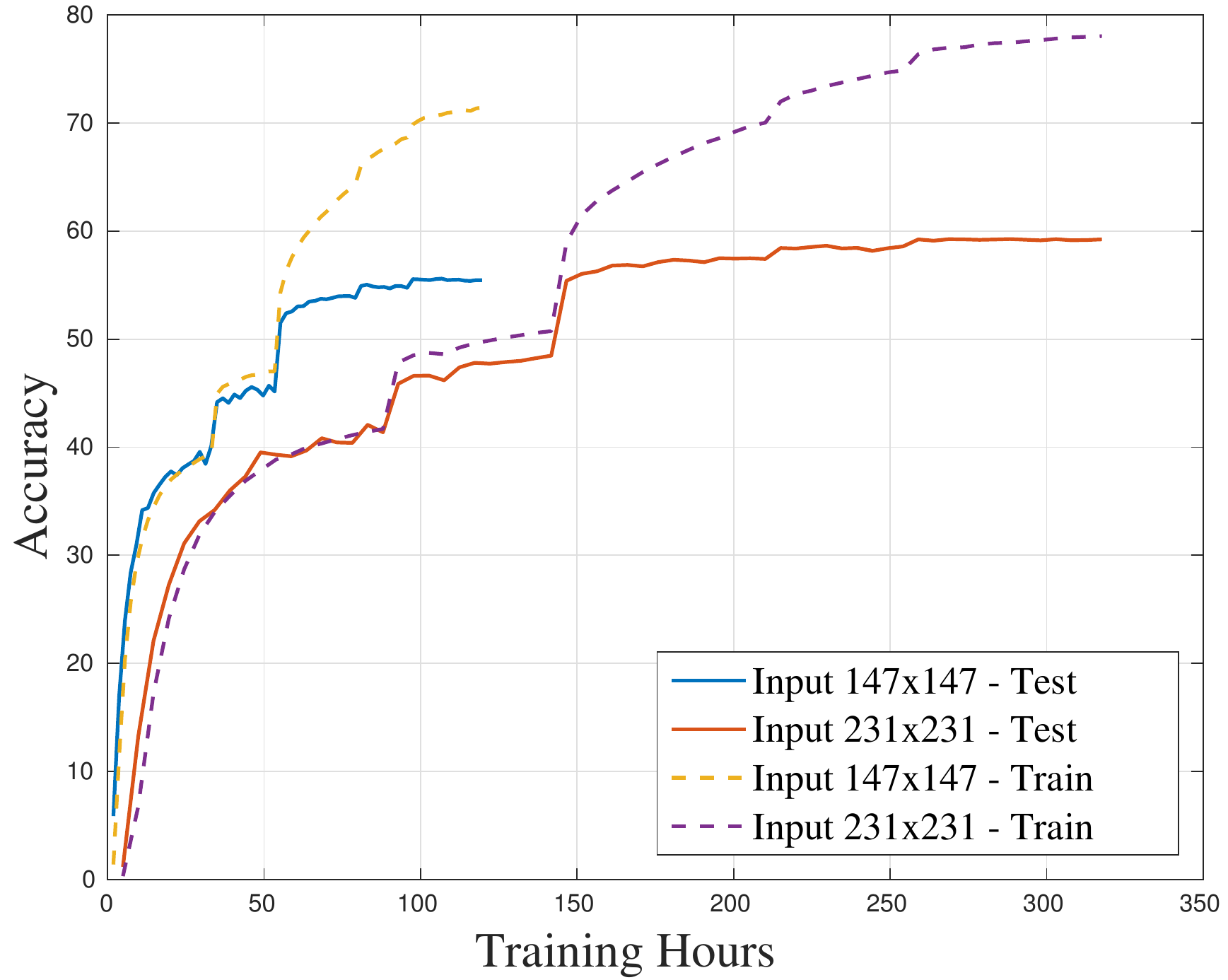}
	\caption{Accuracy as function time obtained using both original OverFeat-\emph{fast} of input resolution $231\times{}231$ and its \emph{pre-train} counterpart having $147\times{}147$ input resolution.}
	\label{fig:originalvsdownsizedhours}
\end{figure}

For the two networks we obtain the train and test accuracies with respect to the number of training epochs and training hours, seen in Figures \ref{fig:originalvsdownsizedepochs} and \ref{fig:originalvsdownsizedhours}. Representing accuracy in terms of epochs allows one to measure how fast the network is learning as data is presented to it, while representation in terms of training time reflects the variable to be optimized.  

We notice in Figure \ref{fig:originalvsdownsizedepochs} that test accuracy stops increasing a few epochs after the last change in learning rate. For this reason, we consider both networks to have been fully trained at the end of 55 epochs resulting in best test accuracies of $59.25\%$ (epoch 55) for the \emph{target} network and $55.55\%$ (epoch 53) for the \emph{pre-train} network. We also observe from Figure \ref{fig:originalvsdownsizedepochs} that, during the first epochs, both test and training accuracies follow the same pattern for the two networks, which suggests that information being learnt by the models is both generalizable and adequate to be represented by the smaller, faster network. 

On the other hand, Figure \ref{fig:originalvsdownsizedhours} highlights the effect of using spatially smaller kernels on training time. OverFeat-\emph{fast} took $269.1$ hours to perform 55 epochs of training while the \emph{pre-train} network only took $106.8$ hours to train on the same amount of data. This reflects a reduction in training time by a factor of $2.51$, which largely agrees with the upper-bounds set by Equation \ref{eq:nummult} and the values of $s_l^2$ in Table \ref{table:conversiontable} for the $3\times{}3\left(2.25\right)$, $5\times{}5\left(2.77\right)$ and $11\times{}11\left(2.49\right)$ kernel resolutions found in the original architecture. 

%

%


\section{Experiments on Pre-training}\label{sec:pretrain}

In this Section we evaluate the effects of rescaling the \emph{pre-train} network at different points in time. Our goal is to maximize the number of epochs trained using the smaller network in order to reduce the overall time necessary for training the network. 

\subsection{Resize-and-Continue Scheduled Training}

Ideally, one would like to be able to fully train a smaller network, upsize its kernels and immediately obtain the test accuracy of the target network. However, as seen in Figure \ref{fig:originalvsdownsizedepochs}, decrease in learning rate and removal of weight decay lead to increase in overfitting, which in turn imposes some constraints to this straightforward approach.

\begin{figure}
	\begin{center}
	\centerline{\includegraphics[width=0.70\columnwidth]{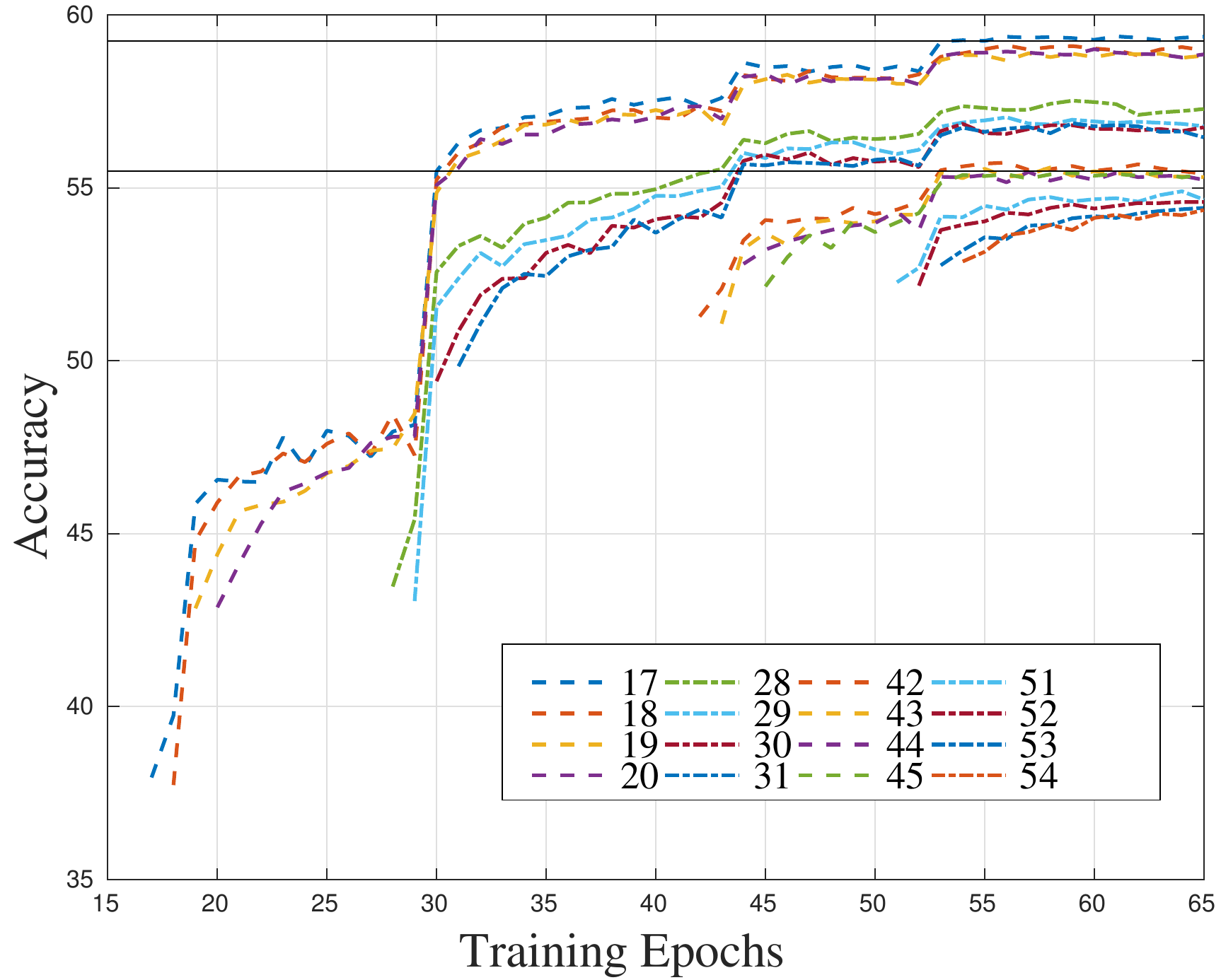}}
	\caption{Effects of rescaling kernels at different epochs. Lower and upper horizontal lines define the maximum accuracies obtained with \emph{pre-train} and \emph{target} networks, respectively.}\label{fig:resizedvcepochs}
	\end{center}
\end{figure}

In this experiment we evaluate the effect of upscaling kernels at different epochs and continuing the scheduled training rule. Since changes in the learning rules had a clear effect on accuracy, we focus on resizing the network before and after these changes. Accuracy curves for each starting epoch are reported in Figure \ref{fig:resizedvcepochs} including threshold lines for the accuracies obtained by the two baseline networks described in Section \ref{sec:preliminary}. Although we still consider a 55 epoch training schedule, the process is carried out until epoch 65 in order to verify possible gains due to continuing training. 

Curves in Figure \ref{fig:resizedvcepochs} reveal some interesting behaviour. Each resized model shows a lower starting accuracy when compared to the $147\times{}147$ input network test curve. This pattern is expected since interpolation will give an imperfect estimate of the desired kernels. On the other hand, the fact that accuracy does not drop too much indicates that knowledge can, at least partially, be transferred using this method.

The same figure also shows a saturation effect. Networks resized at early stages (epochs 17-20) are able to achieve levels of accuracy similar to the $231\times{}231$ input network, while networks resized at late stages (epochs 51-54) can only achieve accuracies below the  $147\times{}147$ input network threshold. Intermediate values of accuracies were obtained when upsizing the \emph{pre-train} network at epochs from 28 to 31, while accuracies close to the one obtained with the $147\times{}147$ input network baseline were obtained by upsizing the pre-train networks at epochs 42-45. Restart training in the vicinity of the last change in learning rate resulted in test accuracies below the \emph{pre-train} network baseline threshold. 

Table \ref{table:resizedaccuracyandtraining} summarizes training times and accuracies for those networks that were able to closely approximate the final accuracy of the \emph{target} network. From this experiment we notice that resizing the \emph{pre-train} network at Epoch 17 produced the same accuracy as the target network even though it takes $49.1$ less hours to finish training, a relative gain of $18.25\%$ in training time. These results show the necessity of upscaling early during training in order to achieve the maximum \emph{target's} accuracy. 

\begin{table}
\caption{Final accuracy and training times for resized networks after a total of 55 epochs. Lower and upper bound accuracies are set by \emph{pre-train} and \emph{target} networks, respectively.}
\label{table:resizedaccuracyandtraining}
    \begin{center}
     	\begin{tabular}{lccc}
			\hline\noalign{\smallskip}
 			Network 	 					&	& Best Accuracy (Epoch) & Total Training Time  \\
			\hline\noalign{\smallskip}
			\emph{Pre-train} (Input 147$\times{}$147)  	&	& 55.55\%(53) 	& \SI{106.8}{\hour}\\
			\emph{Target} (Input 231$\times{}$231) 	&	& 59.25\%(55) 	& \SI{269.1}{\hour}\\
			Resized at Epoch 17 			&	& \textbf{59.25\%}(54) 	& \textbf{220.0 h}\\
			Resized at Epoch 18  			&	& 59.01\%(55) 	& \SI{217.0}{\hour}\\
			Resized at Epoch 19 			&	& 58.84\%(55) 	& \SI{213.9}{\hour}\\
			Resized at Epoch 20  			&	& 58.91\%(54) 	& \SI{210.9}{\hour}\\
			\noalign{\smallskip}\hline
		\end{tabular}
	\end{center}
\end{table}

\subsection{Resize-and-Continue with Extra Training}


It can be observed in Figure \ref{fig:resizedvcepochs} that when resizing from epochs 17, 28, and 42, the subsequent epoch still shows relevant increase in accuracy, which does not happen at the same epochs for the baseline networks since, at those points, test accuracy plateaus, raising the need to change learning rate. 
From this observation we consider maintaining the same learning rule after resizing the networks until there is a drop in test accuracy, from which point on we continue with the predefined learning schedule.  

\begin{figure}
	\centering
	\includegraphics[width=0.70\columnwidth]{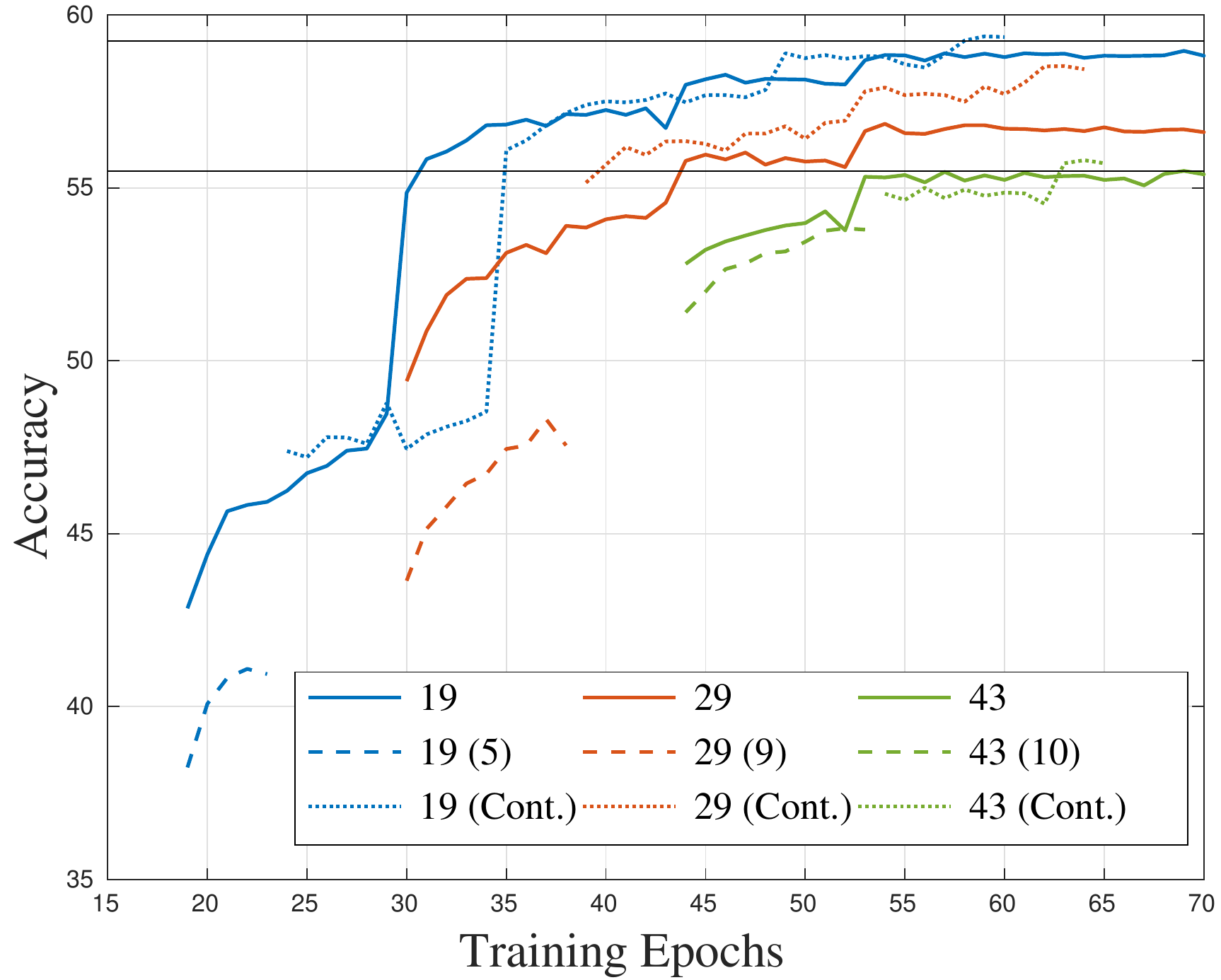} 
	\caption{Accuracy as a function of epochs when training is allowed to continue using current learning rules for a few extra epochs. Learning rule is updated as soon as there is a drop in test accuracy.}
	\label{fig:conttrainepochs}
\end{figure}

\begin{figure}
	\centering
	\includegraphics[width=0.70\columnwidth]{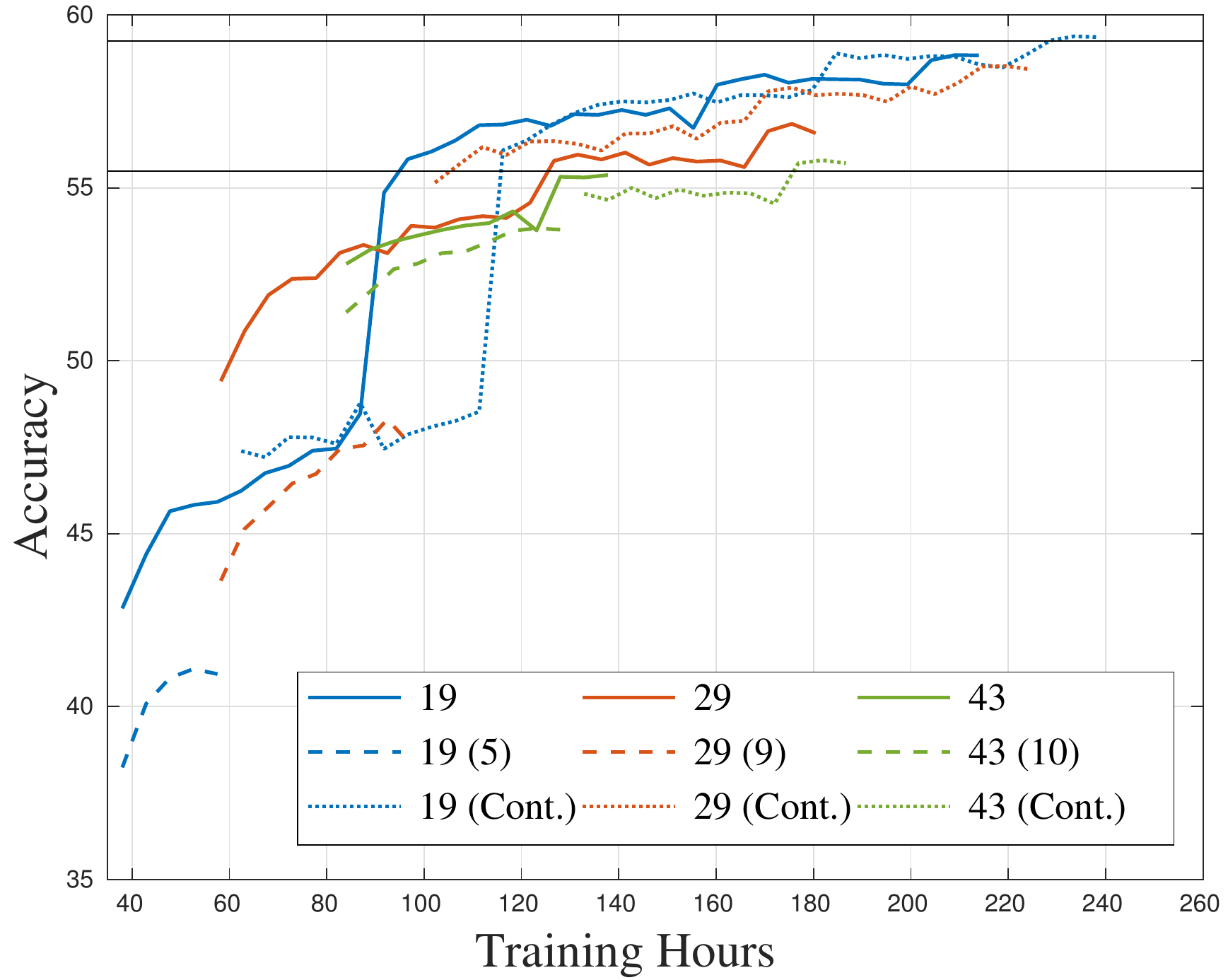}
	\caption{Accuracy as a function of time when training is allowed to continue using current learning rules for a few extra epochs. Learning rule is updated as soon as there is a drop in test accuracy.}
\label{fig:conttrainhours}
\end{figure}

Again we try to maximize the number of epochs run using the \emph{pre-train} network, so we resize and continue the new training procedure at the end of epochs 18, 29 and 43 since these starting points achieved accuracies above the $147\times{}147$ input network threshold during the previous experiment. 
Effects of continuing training using current learning rules for an extra number of epochs are reported in Figures \ref{fig:conttrainepochs} and \ref{fig:conttrainhours} along with the curves produced in the previous experiment for the same restarting points.  

Accuracies and times for both pre-training approaches are reported in Table \ref{table:resizedacontinuecurrent}. For each starting point, it can be seen that training for a number of extra epochs does increase the final accuracy. However, this extra training comes at the cost of slowing the overall training procedure. 

Moreover, we observe that continuing training from Epoch 19 for 5 extra epochs resulted in a test accuracy slightly above the upper-bound defined by the target network in Section \ref{sec:preliminary}. Although the difference is too small to be considered as an actual improvement ($0.11\%$) it does prove that the upper-bound is achievable using the proposed method while avoiding $30.7$ hours of training ($11.41\%$ with respect to the original time).   

\begin{table}
\caption{Best accuracy and total training times for resized networks with extra training.}
\label{table:resizedacontinuecurrent}
    \begin{center}
		\begin{tabular}{lccc}
			\hline\noalign{\smallskip}
 			Network 	 					&	Extra Epochs & Accuracy (Epoch) & Training Time  \\
			\hline\noalign{\smallskip}
			Resized at Epoch 19 (Continued)	&	5 & 59.36\% (59) 	& \SI{238.4}{\hour}\\
			Resized at Epoch 30 (Continued)	&	9 & 58.52\% (64) 	& \SI{224.4}{\hour}\\
			Resized at Epoch 43 (Continued)	&	10 & 55.80\% (64) 	& \SI{186.6}{\hour}\\
			\noalign{\smallskip}\hline
		\end{tabular}
	\end{center}
\end{table}

\subsection{Residual Networks}\label{subsec:resnet}
To prove that our approach can be used on different architectures along with other optimization techniques, we apply our method to the more recent Residual Network \citep{he2015deep} architecture having 34 layers. As suggested by previous results, we resize the \emph{pre-train} network one and two epochs before changing learning rate and verify possible gains in training times. 

For this experiment, training was performed for 90 epochs using mini-batches of 128, weight decay of \num{1e-4}, and momentum equal to \num{0.9}. Learning rate is initially set to \num{1e-1} and it is reduced to \num{1e-2} and \num{1e-3} before epochs 31 and 61. All experiments were run on a single NVIDIA Titan-X using the CuDNN library for FFT based convolutions. Original images crop resolutions were $224\times{}224$ for the \emph{target} network and $160\times{}160$ for \emph{pre-train}.

\begin{figure}
	\centering
	\includegraphics[width=0.70\columnwidth]{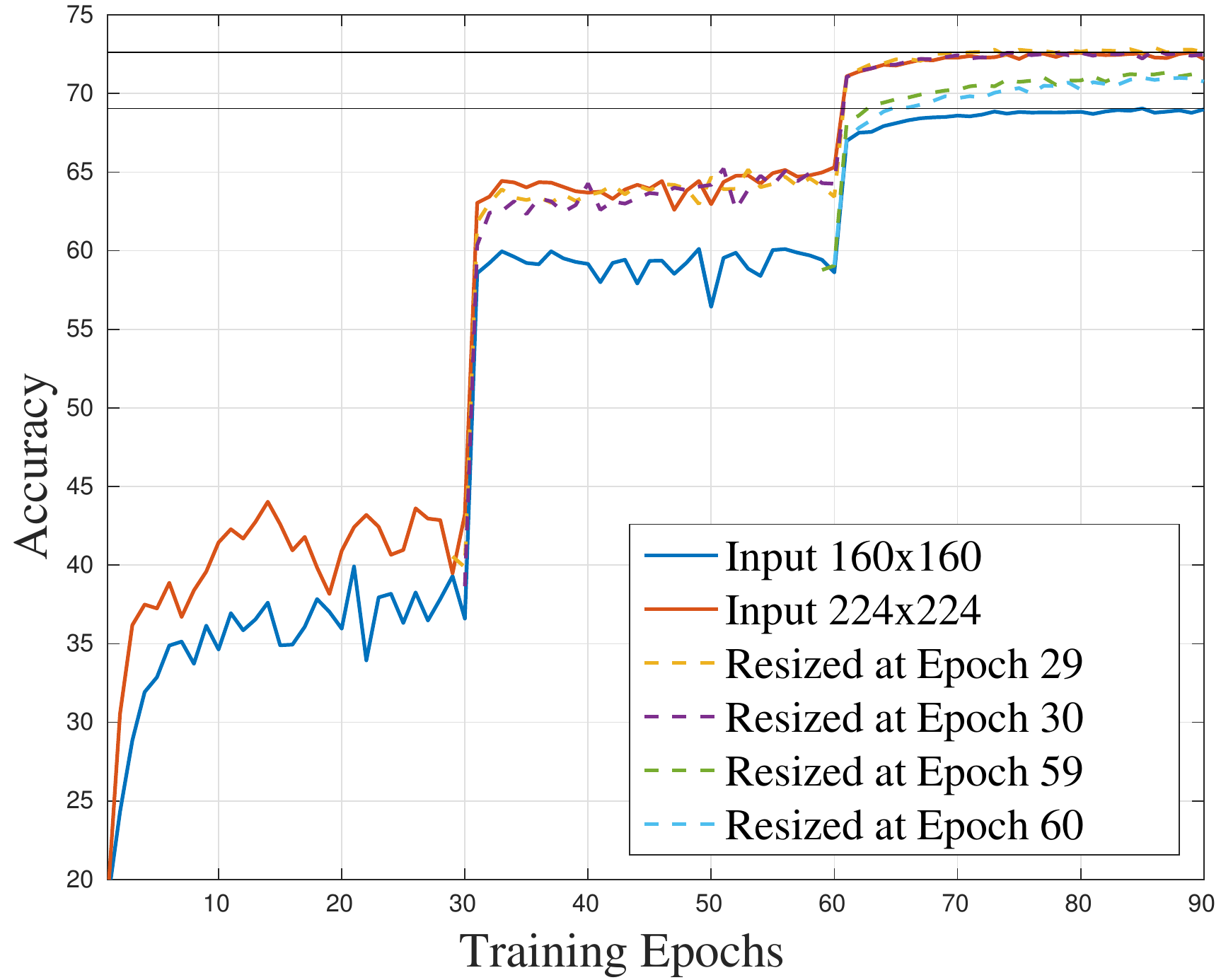}
	\caption{Accuracy curves obtained using ResNet-34 as a function of epochs. Lower and upper horizontal lines define the best accuracies  obtained for the new baseline networks.}
\label{fig:resnetepochs}
\end{figure} 

\begin{figure}
	\centering 
	\includegraphics[width=0.70\linewidth]{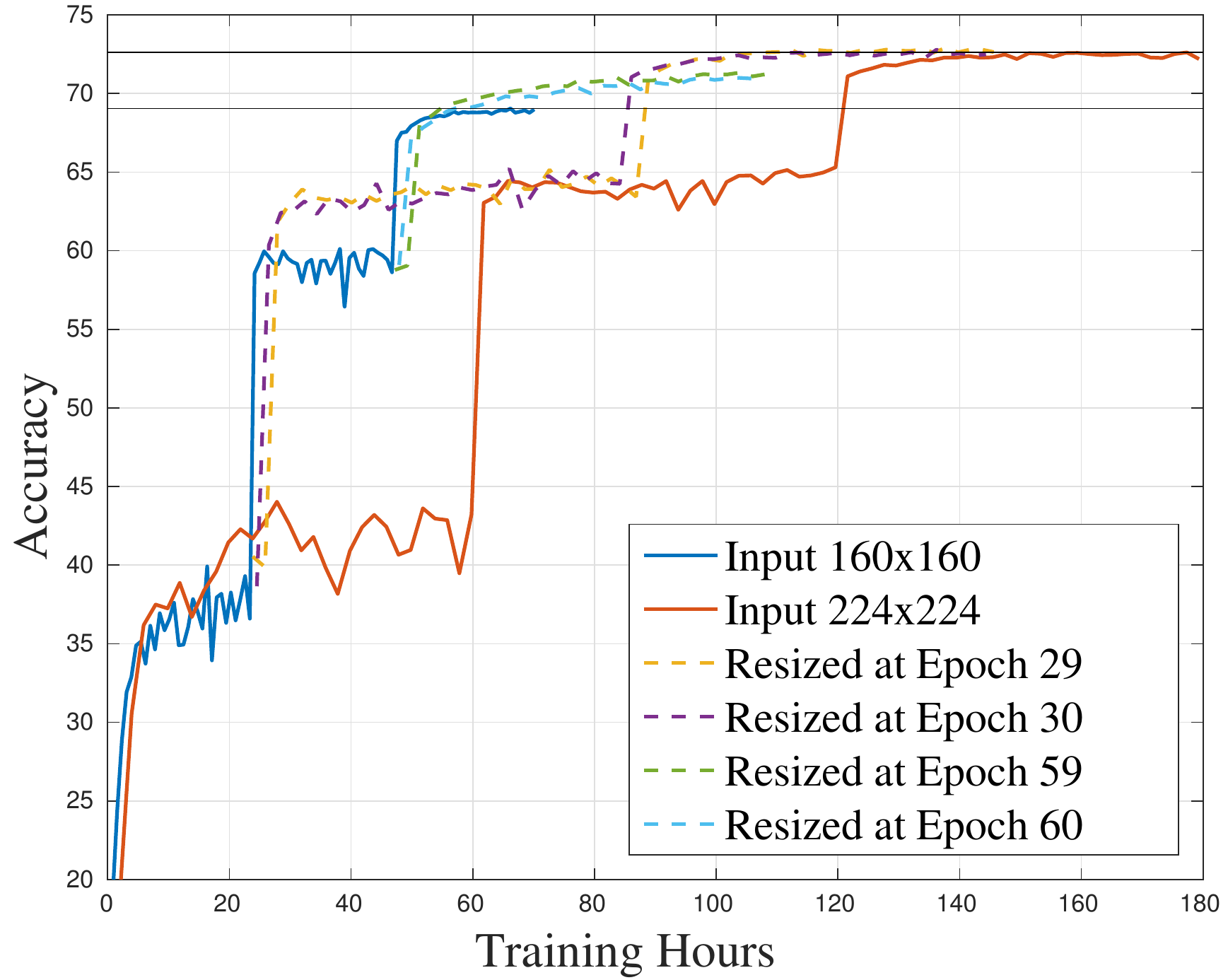}
	\caption{Accuracy curves obtained using ResNet-34 as a function of time. Lower and upper horizontal lines define the best accuracies  obtained for the new baseline networks.}
\label{fig:resnethours}
\end{figure} 

\begin{table}
\caption{Best accuracy and training times for ResNet-34. Training is reduced by $33.7$ hours when upscaling  two epochs before changing learning rate. }
\label{table:resnettraining}
    \begin{center}
		\begin{tabular}{lcc}
			\hline
			\noalign{\smallskip}
 			Network 	 				 & Accuracy (Epoch) & Training Time  \\
			\hline\noalign{\smallskip}
			\emph{Pre-train} ($160\times{}160$)  & 69.05\% (85) 	& \SI{70.09}{\hour}\\
			\emph{Target} ($224\times{}224$) 	 & 72.61\% (89) 	& \SI{179.33}{\hour}\\
			Resized at Epoch 29 	 			 & \textbf{72.91\%} (86) 	& \textbf{145.60 h}\\
			Resized at Epoch 30 	 			 & 72.79\% (86) 	& \SI{144.32}{\hour}\\
			Resized at Epoch 59 				 & 71.36\% (90) 	& \SI{110.02}{\hour}\\
			Resized at Epoch 60 				 & 71.01\% (85) 	& \SI{107.82}{\hour}\\
			\noalign{\smallskip}\hline
		\end{tabular}
	\end{center}
\end{table}

As seen in Figures \ref{fig:resnetepochs} and \ref{fig:resnethours}, resizing at early epochs (29 and 30) allowed the networks to achieve the expected maximum accuracy, while resizing at late epochs (59 and 60) prevented them from doing so. Moreover, when compared to the original architecture, resizing the \emph{pre-train} ResNet at epoch 29 allowed it avoid $33.7$ hours ($18.80\%$) of training and gave in slightly better accuracy.
A summary of these results is reported in Table \ref{table:resnettraining}.

\section{Conclusion}\label{sec:conclusion}

In this work, we have presented a fast way of training CNN that exploits the spatial scaling property of convolutions. Ideally the scaling property would allow a \emph{target} model to be trained from a fully trained \emph{pre-train} network. In practice, however, we have observed that there is an intrinsic saturation process that prevents such n{\"a}ive implementation from succeeding. The longer the \emph{pre-train} network is trained the less likely it is to achieve the performance of the \emph{target} network. Although further investigation is required, to the best of our knowledge this happens because, as the \emph{pre-train} network is trained, the learnt set of weights moves towards a deep local minimum making it difficult to locally find better weights with lower learning-rates. 
 
However, we observe that this effect is mitigated at early stages of learning where testing and training accuracies are similar for both networks. This leads to the conclusion that both networks are learning information that can be generalized, and that can be effectively exploited at both kernel resolutions. This allowed us to use the proposed approach as a pre-training technique where, by resizing the network a couple of epochs before the first scheduled change in learning rate, we were able to obtain the expected \emph{target} accuracy for both OverFeat and ResNet architectures while avoiding $49.1$ hours ($18.25\%$) and $33.7$ hours ($18.80\%$) of training, respectively.   

\bibliographystyle{iclr2017_conference}
\bibliography{pretraincnn}

\end{document}